\documentclass{article}

\usepackage{times}
\usepackage{float}
\usepackage{graphicx} 
\usepackage{subfigure}

\usepackage{amsmath}
\usepackage{amssymb}
\usepackage{natbib}

\usepackage{algorithm}
\usepackage{algorithmic}

\usepackage{hyperref}
\usepackage{multicol}

\usepackage[accepted]{icml2016}

\icmltitlerunning{Object Recognition Based on Amounts of Unlabeled Data}

\begin{document}
\onecolumn
\icmltitle{Object Recognition Based on Amounts of Unlabeled Data}


\icmlauthor{FUQIANG LIU}{fqliu@outlook.com}
\icmladdress{Beijing Institute of Technology, Beijing 100081 CHINA}
\icmlauthor{FUKUN BI}{bifukun@ncut.edu.cn}
\icmladdress{North China University of Technology, Beijing 100144 CHINA}
\icmlauthor{LIANG CHEN}{chenl@bit.edu.cn}
\icmladdress{Beijing Institute of Technology, Beijing 100081 CHINA}

\icmlkeywords{Object Recognition, Boost Picking, Semi-supervised learning}

\vskip 0.3in

\begin{abstract}
This paper proposes a novel semi-supervised method on object recognition. First, based on Boost Picking~\cite{BoostPicking}, a universal algorithm, Boost Picking Teaching (BPT), is proposed to train an effective binary-classifier just using a few labeled data and amounts of unlabeled data. Then, an ensemble strategy is detailed to synthesize multiple BPT-trained binary-classifiers to be a high-performance multi-classifier. The rationality of the strategy is also analyzed in theory. Finally, the proposed method is tested on two databases, CIFAR-10 and CIFAR-100~\cite{cifar}. Using 2\% labeled data and 98\% unlabeled data, the accuracies of the proposed method on the two data sets are 78.39\% and 50.77\% respectively.
\end{abstract}

\section{Introduction}

High-level computer vision applications like detection and recognition have been developing rapidly. Supervised methods~\cite{mohri2012foundations} contribute a lot and representative works include DPM~\cite{Felzenszwalb2010Object}, Alexnet~\cite{krizhevsky2012imagenet}, Highway Network~\cite{highwaynetwork}, Residual Network~\cite{residualnetwork} and so on. Large and open data sets like ImageNet~\cite{fei2010imagenet} and Microsoft COCO~\cite{lin2014microsoft} greatly contribute to this boom. And crowds~\cite{attenberg2011beat,burton2012crowdsourcing}, who labeled these data, also play an important role in the development.

But the size of unlabeled data in practice is much larger than that of labeled data and it is too expensive to label a large number of data by humans, especially the data are generated all the time. Humans do not need so many labeled data like the computer to learn a new knowledge or concept. A baby gradually knows the world from his practices and parent's guide. In this process, unlabeled data might be more important. As a result, this paper researches in how to mainly use unlabeled data to teach computer recognition.

To begin with, this paper presents the theoretical basis (see section~\ref{sec:theoretical basis}) of the proposed method. Kalal's works~\cite{kalal2010pn,kalal2012tracking} presented P-N Learning to update object models in tracking application. Kalal proved that P-N experts can train a strong classifier only if eigenvalues of the transforation matrix are all smaller than 1 (equation~\ref{equ:reconstruct iteration}). Liu's work~\cite{BoostPicking} pointed out that P-N Learning ignored the bias of supervised models in training set and proposed Boost Picking to convert supervised classification to semi-supervised classification. Using Boost Picking, two weak classifiers could train a strong classifier mainly by unlabeled data. Liu derived that Boost Picking could work effectively even taking the bias into account. Furthermore, Liu theoretically proved that Boost Picking would train a supervised classification model mainly by un-labeled data as effectively as the same model trained by all labeled data, only if recalls of the two weak classifiers are both greater than 0 and sum of their precisions is greater than 1 (equation~\ref{equ:condition}).

And then, referring to Boost Picking, we design a method named "Boost Picking Teaching (BPT)" for binary-classification (see section~\ref{sec:bpt}). BPT includes a binary-classifier(named BPT-trained binary-classifier) and two "teachers". Like a baby knowing the world, BPT improves the performance of the binary-classifier by practicing with unlabeled data and teachers' guides. Different from Boost Picking, BPT does not need any labeled data to initialize the binary-classifier, which is like that a baby knowing the world originally from unlabeled data. BPT achieves that a strong binary-classifier could be trained totally by unlabeled data under a specific and loose condition. The condition is as same as that in Boost Picking~\cite{BoostPicking} (equation~\ref{equ:condition}). The binary-classifier classifies those unlabeled data and then "teachers" find the errors and correct the classifier by retraining. This step is like bootstrap~\cite{Sung1995Example}, but BPT uses estimated labels rather than true labels. Liu's work theoretically guarantees that even though the "teachers" are not good enough, the binary-classifier could be trained to be a strong one just by repeating the above step. However, in the recognition application (see section~\ref{sec:recogition implementation}), we still use a few labeled data to train qualified "teachers". Under the relatively complicated recognition circumstance, it would be a little bit difficult to make sure the "teachers" satisfy the above condition (equation~\ref{equ:condition}), if the "teachers" are designed based on un-supervised methods.

Furthermore, to achieve recognition, this paper proposed a high-performance multi-classifier based on BPT-trained binary-classifiers (see section~\ref{sec:multi-bpt}). First, we briefly discuss why it is essential to combine multiple BPT-trained binary-classifiers rather than directly train a multi-classifier like BPT. Then, we details the framework of the ensemble classifier (named BPT-multi-classifier) as well as the synthesizer (ensemble strategy). A novel idea is that the synthesizer is composed by BPT-binary classifiers. It is also theoretically analyzed that the framework of BPT-multi-classifier is reasonable.

Based on BPT-multi-classifier, this paper proposes a practical implementation on object recognition (see section~\ref{sec:recogition implementation}). The basic classification model of BPT-trained binary classifier is a support vector machine(SVM)~\cite{suykens1999least}. And K-means~\cite{Coates2012Learning} is adopted to extract features from the images. As for the "teachers" in BPT, they are achieved by logistic regression models~\cite{hosmer2004applied} that are trained by a few labeled data. And features that "teachers" use are extracted by Histogram of Oriented Gradient (HOG)~\cite{Dalal2005Histograms}. CIFAR-10 and CIFAR-100 ~\cite{cifar} are involved in the experiment (see section~\ref{sec:experiment}) to test the effectiveness of the proposed method. The accuracies of the proposed method on the two databases are 78.39\% and 50.77\% respectively. In this test, the proposed method use only 2\% labeled data and 98\% unlabeled data (the size of the two data sets are both 60000).

The BPT-multi-classifier could be regarded as a semi-supervised method because labeled data and unlabeled data are both used. But different from previous object recognition works based on semi-supervised methods~\cite{Chen2013NEIL,Rosenberg2005Semi,Fergus2009Semi,Prest2012Learning}, BPT-multi-classifier does not involve any web data. And unlike one-shot learning~\cite{Li2006One,Socher2013Zero} and baby learning~\cite{Liang2014Computational}, the proposed work does not use any prior knowledge or pre-trained model. In fact, the thought of the proposed method is different from the thoughts of previous semi-supervised methods. The basic units of BPT-multi-classifier are some traditional supervised models (SVM and logistic regression). The key idea of our work is that weak classifiers could train a strong classifier just using unlabeled data and their estimated labels (the weak classifiers may be trained by a few labeled data).

\section{Theoretical Basis}
\label{sec:theoretical basis}
This section presents the theoretical basis of BPT.

In~\cite{kalal2012tracking}, new forms of the tracked object are updated by P-N Learning. P-N Learning adopts P-N experts to estimate errors of the detector. P-expert finds the false negatives from the negative outputs of the detector and N-expert finds the false positives from the positive outputs. Kalal theoretically proved that P-N Learning could train a detector whose error converges into zero. Define $\overrightarrow{e}(k)$ as the errors in kth iteration,
\begin{equation} \label{equ:error}
\overrightarrow{e}(k)=[\alpha(k), \beta(k)]^{T}
\end{equation}
where $\alpha(k)$ represents false positives and $\beta(k)$ represents false negatives in outputs.

Define $P^{+}$,$R^{+}$ as the precision and the recall of P-expert respectively. Similarly, define $P^{-}$,$R^{-}$ as the precision and the recall of N-expert. The recursive equations are as equation~(\ref{equ:iteration})\cite{kalal2012tracking}.
\begin{equation}
\label{equ:iteration}
\left\{
\begin{aligned}
\alpha(k+1)=\alpha(k)-R^{-}\cdot\alpha(k)+\frac{1-P^{+}}{P^{+}}R^{+}\cdot\beta(k)\\
\beta(k+1)=\beta(k)-R^{+}\cdot\beta(k)+\frac{1-P^{-}}{P^{-}}R^{-}\cdot\alpha(k)
\end{aligned}
\right.
\end{equation}

Reconstruct equation~(\ref{equ:error}) and~(\ref{equ:iteration}),
\label{equ:reconstruct iteration}
\begin{equation}
\begin{array}{cc}
\overrightarrow{e}(k+1)=M\overrightarrow{e}(k)\\
M=[\begin{array}{cc}
1-R^{-} & \frac{1-P^{+}}{P^{+}}R^{+}\\
\frac{1-P^{-}}{P^{-}}R^{-} & 1-R^{+}\\
\end{array}]\\
\end{array}
\end{equation}

Assuming $\lambda_1$, $\lambda_2$ are eigenvalues of transformation matrix $M$, Kalal concludes that $\overrightarrow{e}$ will converge to zero only if $\lambda_1$, $\lambda_2$ are all smaller than 1~\cite{kalal2010pn}. In practice, the classifier almost cannot be trained to be errorless even through P-N experts could find out all false positives and false negatives, because bias~\cite{geman1992neural,james2003variance} always exits.

Supposing $x$ is an example from a feature-space $X$ and $y$ is its corresponding label from a label space $Y=\{1,-1\}$, $X_l$ is a labeled data set with its label set $L_l$, and $X_u$ is an unlabeled data set. Supposing that $L_u$ are the in-existent true labels of $X_u$, define $F:x\rightarrow y$ as an supervised learning model trained by 100\% labeled data set $X_{(100\%)}=\{X_l, X_u\}$ with $L_{(100\%)}=\{L_l, L_u\}$. Boost Picking~\cite{BoostPicking} aims to train an effective model $f\rightarrow F$ by $X_l$, $L_l$ and $X_u$. In Boost Picking~\cite{BoostPicking}, two weak classifiers pick out the false positives and false negatives from the outputs of $f$ just like what P-N experts do. Then put these estimated errors into the re-training set with their estimated labels and retrain $f$ like bootstrap~\cite{Sung1995Example} (bootstrap uses true labels but Boost Picking use the estimated labels). Liu proved that $f\rightarrow F$ in the iteration if equation~\ref{equ:condition} is satisfied, both in theory and experiment. Boost Picking could still work even considering the bias.

\begin{equation}\label{equ:condition}
\left\{
\begin{aligned}
R^{+}R^{-}\neq0\\
P^{+}+P^{-}>1
\end{aligned}
\right.
\end{equation}

Equation~\ref{equ:condition} guides how to use two weak classifiers to train a strong binary classifier. Based on this conclusion and referring to Boost Picking, this paper proposes a modification, Boost Picking Teaching (BPT), for binary classification. And furthermore, BPT is expanded to multiple-classification and a complicated application, object recognition.

Boost Picking is to train a strong binary classifier by labeled data $X_l$ and unlabeled data $X_u$, while BPT is to train a strong binary classifier by unlabeled data $X_u$. Unlike Boost Picking, BPT does not need labeled data to initialize the binary classifier. Because the initial set and re-training set of Boost Picking rely on a certain number of labeled data, the classifier trained by Boost Picking would be over-fitting if labeled data are not enough. In fact, Boost Picking needs 25\% labeled data as well as 75\% unlabeled data to train a strong classifier. But in theory, BPT does not need labeled data. It should be noted that a few labeled data might be used in BPT to train qualified "teachers". Because the test sets in~\cite{BoostPicking} are relatively simple and small, Boost Picking just used an unsupervised method to train the two weak classifiers, FP and FN Finders. But the proposed method is tested on large databases in this paper, and this is why the two weak classifiers, "teachers", in BPT need a few labeled data to train themselves.

\section{Boost Picking Teaching for Binary Classification}
\label{sec:bpt}
\begin{figure*}[htb]
\vskip -0.15in
\includegraphics[width=\textwidth]{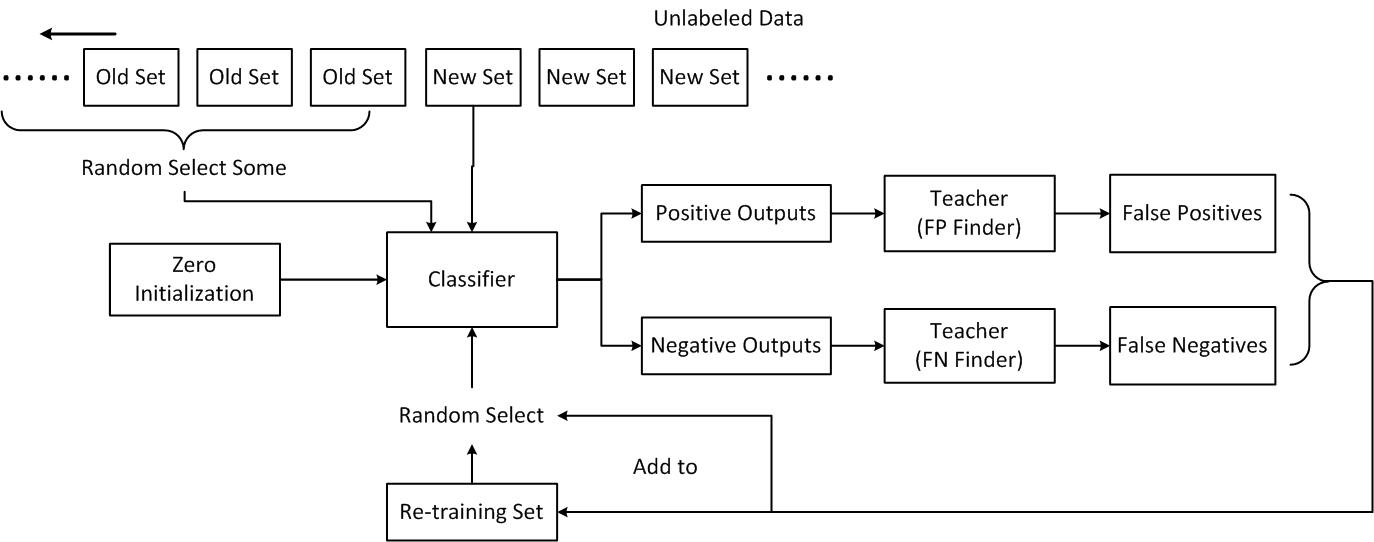}
\caption{Framework of BPT for Binary Classification}
\label{fig:framework}
\vskip -0.15in
\end{figure*}

This section details BPT for binary classification. BPT could train a strong classification $f$ just using unlabeled data $X_u$ in theory. The framework of BPT is showed as Fig~\ref{fig:framework}. Though it refers to the framework of Boost Picking~\cite{BoostPicking}, there are some modifications to achieve the goal, training a supervised binary-classifier without labeled data.

Supposing $f$ is the binary classifier parameterized by $\Theta$. The first step of BPT is that $\Theta$ are initialized as $\textbf{0}$. Actually, there is no key difference between zero initialization and random initialization in BPT.

Second, one part of unlabeled data are classified by the binary classifier. And two "teachers", FP Finder and FN Finder, pick out false positives $FP$ and false negatives $FN$ in the outputs. Errors of the two "teachers" are allowed, only if the "teachers" satisfy equation~\ref{equ:condition}. Then, these estimated errors with their amended labels(the labels of examples in $FP$ are set as -1 and the labels of examples in $FN$ are set as 1) are put into the re-training set $X_R$, $X_R=X_R\bigcup\{FP,FN\}$. And the classifier $f$ is re-trained by $X_r$, where $X_r \subset X_R$ and $\{FP,FN\}\subset X_r$. Next, go back to the beginning of this step and repeat classifying, correcting and re-training. The iteration process stops when the performance of the classifier is stable. At this time, the classifier is trained to be parameterized as $theta_S$. In every generation, only part of examples in $X_R$ are used in re-training because $X_R$ would gradually grow to be a large set in the iteration proceeding. $X_r=randomselect(X_R)\bigcup\{FP,FN\}$. Considering efficiency, examples from $\{FP,FN\}$ are all involves in $X_r$. To avoid being over-fitting, $X_r$ is generated by randomly selecting $X_R$. The idea that put mis-classified examples into training set resembles bootstrap~\cite{Sung1995Example}. But unlike bootstrap, BPT does not need the true labels of the data.

Third, a new part of unlabeled data are classified by the classifier. The next operations are almost similar with the second step. But there are two special operations. One is that the classifier is parameterized as $\theta_S$ initially. Another one is that some examples randomly selected from previous parts are also classified by the binary classifier like Fig~\ref{fig:framework}. Repeat this step until no more data to input. If we do not take the labeled data might used to train "teachers" into account, BPT could train an effective binary-classifier just using un-labeled data, only if the "teachers" satisfy equation~\ref{equ:condition}.

\vskip -0.15in
\begin{algorithm}[htb]
   \caption{Boost Picking Teaching (Binary Classification)}
   \label{code:example}
\begin{algorithmic}
\begin{multicols}{2}
   \STATE Initialization:
   \STATE $\Theta_S = \textbf{0}$
   \STATE $X_R =\varnothing$, $X_P =\varnothing$
   \REPEAT
   \STATE {\bfseries Input:} un-labeled data $X_{u_i}$
   \STATE $\Theta_s=\Theta_S$
   \STATE $X_P=X_P \bigcup \{X_{u_{i-1}}\}$
   \STATE $Now=f(X_{u_i|\Theta_s})$
   \REPEAT
   \STATE $Pre=Now$
   \STATE $V=f(ransel(X_P)|\Theta_s$)
   \STATE $\{FP,FN\}=Teachers(\{Pre,V\})$
   \STATE $X_R = X_R\bigcup \{FP,FN\}$
   \STATE $X_r=ransel(X_R)\bigcup\{FP,FN\}$
   \STATE $\Theta_s=ReTrain(f,X_r,\Theta_S)$
   \STATE $Now=f(X_{u_i}|\Theta_s)$
   \UNTIL $\sum|Pre-Now|<threshold$
   \STATE $\Theta_S=\Theta_s$
   \UNTIL{$noInput$}
\end{multicols}
\end{algorithmic}
\end{algorithm}
\vskip -0.15in

For Boost Picking~\cite{BoostPicking}, the training set includes the labeled data in every generation, which causes that the classifier relies on those labeled data too much. BPT adopts random selection to solve this problem but random selection increases the computational complexity of BPT. BPT inputs unlabeled separately, which is convenient to expand training set and involve new data. But BPT is not a critical online learning~\cite{Shalev2012Online,Bertsekas2015Incremental,Brooms2006Stochastic} method because BPT does not abandon any previous data. The pseudocode of BPT is shown in algorithm~\ref{code:example}. $ransel$ represents "Random Select". The convergence condition is that the classifier is stable in the current part of unlabeled data. It is set empirically because the binary classifier would converge more quickly compared to the condition that the classifier should be stable in the whole unlabeled data.

\begin{figure*}[htb]
\centering
\subfigure[accuracy]{\includegraphics[width=0.23\textwidth]{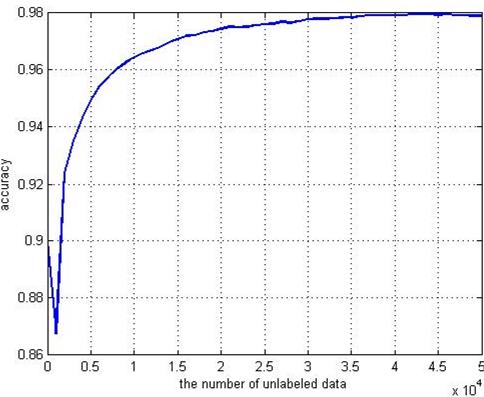}}
\subfigure[precision]{\includegraphics[width=0.23\textwidth]{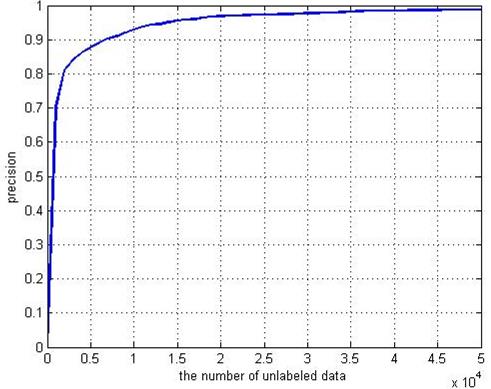}}
\subfigure[recall]{\includegraphics[width=0.23\textwidth]{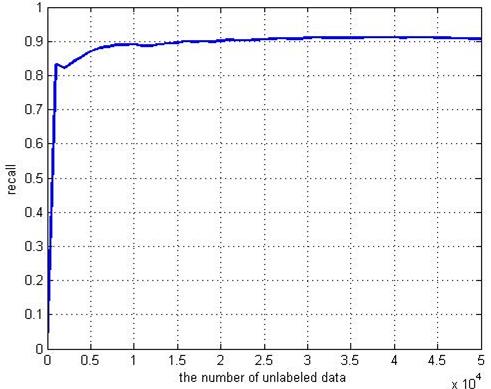}}
\subfigure[f1]{\includegraphics[width=0.23\textwidth]{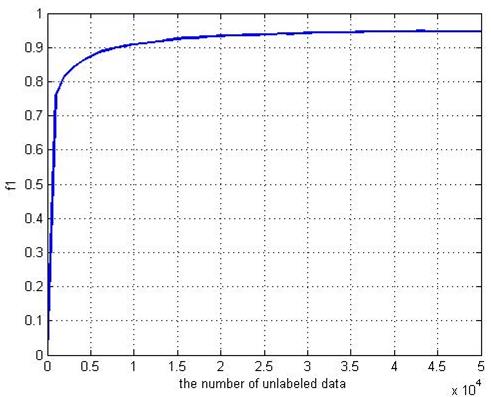}}
\caption{The performance of BPT with synthetic "teachers".}
\label{fig:synthetic}
\end{figure*}

Fig~\ref{fig:synthetic} shows the performance of BPT with synthetic "teachers" whose precisions and recalls are specific. The implementation of the binary classifier for recognition is declared in section~\ref{sec:recogition implementation}. Examples form one class are regarded as positives, the others are regarded as negatives. CIFAR-10~\cite{cifar} is used in this ideal experiment. Recalls and Precisions of the synthetic "teachers" are 0.6. We compute the accuracy, recall, precision and $F1=\frac{2\cdot P\cdot N}{P+N}$ of the binary-classifier in training set(100\% unlabeled data). When the unlabeled data have not yet input, the binary-classifier is parameterized as $\textbf{0}$. Because the data includes 10 classes and the number of every class is almost equal, the accuracy of the classifier is approximately 0.9 initially. Recall, precision and $F1$ are all 0 at the beginning. According to Fig~\ref{fig:synthetic}, all these indexes increase with the proceeding of BPT (with synthetic "teachers"). This ideal experiment proved that BPT could train an effective binary-classifier just using un-labeled. When the "teachers" are not easy to trained to satisfy equation~\ref{equ:condition}, a few labeled data might be used to train qualified "teachers". And in this time, BPT becomes a semi-supervised method.

\section{Multi-Classification Based on BPT}
\label{sec:multi-bpt}
This section details how to construct a multi-classifier based on the BPT-trained binary-classifiers. First, the reason of converting multi-classifier to multi binary-classifiers is briefly proposed. Second, this part presents the multi binary-classifiers. Third, the ensemble strategy on combining these binary-classifiers to be a multi-classifier is detailed.
\subsection{Why Not Directly Train a Multi-Classifier}
Suppose that $f'$ is a multi-classifier, $x$ is an example from a feature-space X and $y$ is its corresponding label from a label space $Y=\{1,2,\cdots,N\}$, where N is the number of the class. This part discusses whether $f'$ could be directly guided by "teachers".

Assume that $False I$ means the examples which are mis-classified as "I" by $f'$, where $I\in\{1,2,\cdots,N\}$. Like BPT, there should be a "teacher" to find out the $False I$. Suppose that $T_i$ is the "teacher" who finds "i" examples from $\{x|f'(x)\neq i\}$, where $i\in\{1,2,\cdots,N\}$. Then assuming that $TeacherI$ picks out $False I$ from $\{x|f'(x)=I\}$, $TeacherI=ensemble\{T_i|i\neq I\ and\ i\in\{1,2,\cdots,N\}\}$. Assuming $P_i$ is the precision of $T_i$ and $R_i$ is the precision of $T_i$, resembling equation~\ref{equ:iteration}, the recursive equation of the errors of $f'$ are written as equation~\ref{equ:multi error}.


\begin{equation}
\label{equ:multi error}
\left\{
\begin{aligned}
e_1 (k+1)=e_1(k)-\sum\limits_{i\neq1} R_i\cdot a_{i,1}\cdot e_1(k)+ \sum\limits_{j\neq1}\frac{1-P_1}{P_1}R_1\cdot e_j(k)\\
e_2 (k+1)=e_2(k)-\sum\limits_{i\neq2} R_i\cdot a_{i,2}\cdot e_2(k)+ \sum\limits_{j\neq2}\frac{1-P_2}{P_2}R_2\cdot e_j(k)\\
\vdots  \qquad \qquad \qquad \qquad \qquad \vdots  \qquad \qquad \qquad \qquad \qquad   \vdots \qquad \\
e_N (k+1)=e_N(k)-\sum\limits_{i\neq N} R_i\cdot a_{i,N}\cdot e_N(k)+ \sum\limits_{j\neq N}\frac{1-P_N}{P_N}R_N\cdot e_j(k)\\
\end{aligned}
\right.
\end{equation}
where $e_i$ represents $False I$($i=I$). Supposing $l(x)$ represents the true class of $x$, $a_{i,j}$ represents the proportion between $\{x|f'(x)=j\ and\ l(x)=i\}$ and {x|f'(x)=j}, where $i\neq j$. And it is obvious that $\forall\,j\in\{1,2,\cdots,N\}$, $\sum\limits_{i\neq j}a_{ij}=1$. Assume $\overrightarrow{e}=\{e_1,e_2,\cdots,e_N\}$. Like equation~\ref{equ:reconstruct iteration}, equation~\ref{equ:multi error} could be reconstructed to be equation~\ref{equ:multi error reconstruct}.

\begin{equation}
\label{equ:multi error reconstruct}
\overrightarrow{e}(k+1)=M\overrightarrow{e}(k)
\end{equation}

The transformation matrix $M$ is a matrix whose size is $N\times N$. $M(i,j)=1-\sum\limits_{i\neq j}R_i\cdot a_{i,j}\cdot e_j(k)$ when $i=j$ and $M(i,j)=\frac{1-P_i}{P_i}R_i\cdot e_j(k)$ when $i\neq j$. Based on the well founded theory of dynamic system~\cite{robustandoptimalcontrol}, if the absolute values of eigenvalues of $M$ are all less than 1, the error of $f'$ would converge to 0; otherwise the error would diffuse.

Here, simplify $M$ to be easily analyze. Assume that $\forall\,i,j\in\{1,2,\cdots,N\}\ and\ i\neq j,\ \ a_{i,j}=\frac{1}{N-1}$. This assumption means that the true labels of the examples in $False I$ might be any other class in equal probability. Furthermore, suppose that the precision and recall of every $T_i$ are equal, $\forall\, i,\ P_i=P,\ and \ R_i=R$, where $P,\ R\in [0,\ 1]$. Suppose that $\lambda_{i}$ is the eigenvalue of the transformation matrix $M$. The expressions of $\lambda_{i}$ is equation~\ref{equ:eigenvalue}.

\begin{equation}
\label{equ:eigenvalue}
\left\{
\begin{aligned}
\lambda_1,\cdots, \lambda_{N-1}=\frac{P-R}{P}\\
\lambda_N = \frac{P+(N-1)\cdot R-N\cdot P \cdot R}{P}
\end{aligned}
\right.
\end{equation}

It is obvious that
\begin{equation}\label{equ:condition multi}
\begin{aligned}
\max(|\lambda|)<1,\ when\ P>\frac{N-1}{N}\\
\max(|\lambda|)=1,\ when\ P=\frac{N-1}{N}\\
\max(|\lambda|)>1,\ when\ P<\frac{N-1}{N}
\end{aligned}
\end{equation}
no matter what $R$ is. It is concluded that if the number of class $N$ increases, $P$ must increase to make sure that $\max(|\lambda|)$ less than 1, which means the recursive equation~\ref{equ:multi error reconstruct} could converge to 0 in the iteration proceeding.


\begin{figure*}[htb]
\centering
\vskip -0.15in
\subfigure[]{\includegraphics[width=0.4\textwidth]{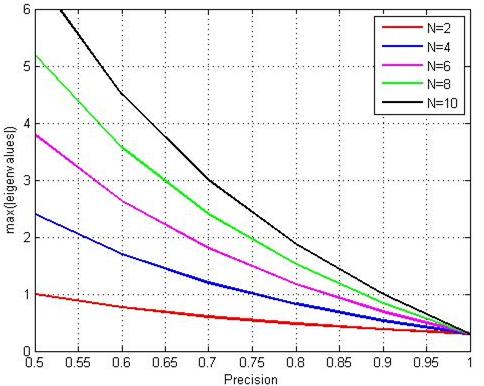}}
\subfigure[]{\includegraphics[width=0.4\textwidth]{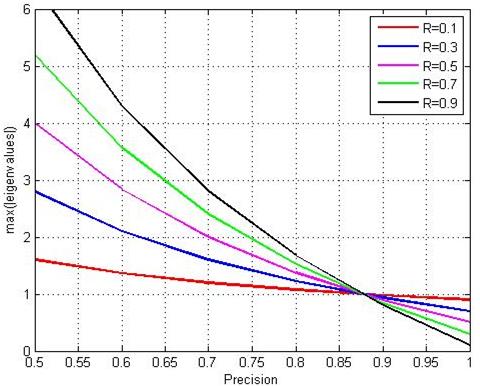}}
\caption{Analysis on the relationship between the maximum absolute value of eigenvalues and $(P,\,R,\,N)$.}
\label{fig:eigvalue}
\vskip -0.15in
\end{figure*}

To better illustrate the relationship between the maximun absolute value of these eigenvalues and $(Recall,\ N)$, we simulate the eigenvalues in various conditions and Fig~\ref{fig:eigvalue} (a-b) are the simulated results. In Fig~\ref{fig:eigvalue} (a), recall $R$ is set as 0.6 constantly. (a) shows that the precision need to increase to guarantee that errors of the binary-classifier converge to 0, which is as same as the conclusion based on equation~\ref{equ:condition multi}. In Fig~\ref{fig:eigvalue} (b), $N$ is set as 8 constantly and $R$ is varied. (b) shows that if the maximum absolute value of eigenvalues is 1, the precision would be certain no matter what $R$ is. This conclusion also matches equation~\ref{equ:condition multi}.

Based on equation~\ref{equ:condition multi} and Fig~\ref{fig:eigvalue}, the condition of "teachers" would be rigorous if training a multi-classifier using unlabeled data like BPT, especially the number of class $N$ is large. Compared to equation~\ref{equ:condition}, the above condition~\ref{equ:condition multi} does not mean that the "teachers" are weak classifiers anymore. As a result, directly training a effective qualified multi-classifier using unlabeled data like BPT is relatively hard. It is better to combine multiple BPT-trained binary-classifiers to be an effective multi-classifier.

\subsection{From Multi Classification to Multi Binary-Classification}
The multi-classification problem is separated to multiple $1\,vs\,N-1$ binary-classification problems, and then synthesize results of the multiple binary-classifiers. This part details the multiple binary-classifiers.

Assuming $f_1,\,f_2,\,\cdots,\,f_N$ are the binary-classifiers used to compose the multi-classifier, each classifier is trained by BPT using unlabeled data(see section~\ref{sec:bpt}). It should be noted that BPT might need a few labeled data to train qualified "teachers", especially when the database is complicated and large. $f_i,\ i\in \{1,\,2,\,\cdots,\,N\}$ is to find out examples those belong to $i$ class from all data.

\begin{table*}
\vskip -0.1in
\begin{center}
\caption{Performance of each binary-classifier trained by BPT with synthetic "teachers" whose precisions and recalls are 0.6}
\label{table:binary classifier for multi}
\begin{tabular}{lllllllllll}
\hline\noalign{\smallskip}
& $f_1$ & $f_2$ & $f_3$ & $f_4$ & $f_5$ & $f_6$ & $f_7$ & $f_8$ & $f_9$ & $f_{10}$\\
\noalign{\smallskip}
\hline
\noalign{\smallskip}
accuracy\ \ \ \ \ & 0.982& 0.988& 0.973& 0.968& 0.977& 0.976& 0.985& 0.980& 0.989& 0.987\\
precision & 0.974& 0.985& 0.967& 0.956& 0.969& 0.964& 0.978& 0.978& 0.987& 0.984\\
recall & 0.934& 0.953& 0.885& 0.858& 0.910& 0.909& 0.952& 0.926& 0.959& 0.951\\
$F1$ & 0.953& 0.968& 0.924& 0.904& 0.938& 0.936& 0.965& 0.951& 0.973& 0.968\\
\hline
\end{tabular}
\end{center}
\vskip -0.1in
\end{table*}
\setlength{\tabcolsep}{3pt}

There are another ideal experiment to show the performances of these binary-classifiers. As the ideal experiment in section~\ref{sec:bpt}, the "teachers" in BPT are synthetic and their precision and recalls are all 0.6. We use CIFAR-10~\cite{cifar} as the experimental data and the specific implementation (including feature extraction and supervised learning model) of the binary-classifier is proposed in section~\ref{sec:recogition implementation}. Table~\ref{table:binary classifier for multi} shows the accuracy, precision, recall and $F1$ of each BPT-trained binary-classifier in training data(100\% unlabeled data). Under ideal condition, these binary-classifiers could be trained very well.

\subsection{Ensemble Strategy}
\label{subect:ensemble}
\begin{figure*}[htb]
\begin{center}
\includegraphics[width=\textwidth]{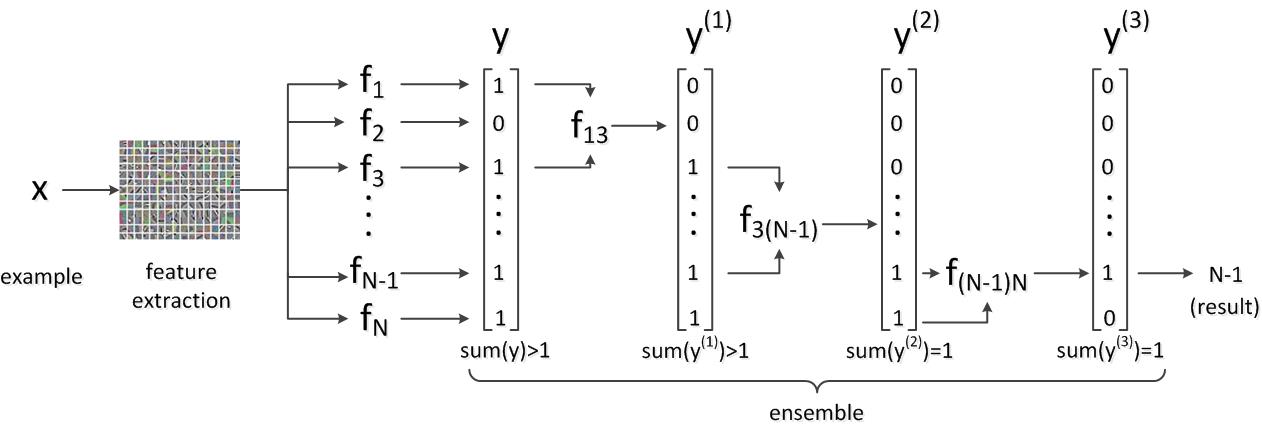}
\caption{the pipeline of the multi-classifier}
\label{fig:pipeline}
\end{center}
\end{figure*}

This section introduces how to synthesize multiple BPT-trained binary-classifiers to be a high-performance multi-classifier. The main reason of ensemble is that there might be more than one binary-classifiers classify a same one example as a positive output, which means one example might be classified to be multiple classes. Fig~\ref{fig:pipeline} is the pipeline of the multi-classifier including both multiple PBT-trained binary-classifiers and the ensemble strategy.

Supposing the data contains examples from $N$ classes, $N$ binary-classifiers ($1\,vs\,(N-1)$) are train well by BPT (see section~\ref{sec:bpt}). Assuming $x$ is an example from the data set, $y = \{f_1{x},\,f_2{x},\,\cdots,\,f_N{x}\}'$ and it is possible that $sum(y)\geq1$. After the ensemble module, there should be $sum(y)=1$. The ensemble module is composed by $frac{N(N-1)}{2}$ binary-classifiers. Supposing $f_{ij}$ is a binary-classifier that is trained by $X_ij={x\in X_u|f_i(x)+ f_j(x)=1\,and\,i\neq j}$ using BPT, $f_{ij}$ is aimed to determine whether $x$ belongs to class $i$. The example $x$ is input into the $N$ binary-classifiers and the output is a $N\times 1$ vector $y$. $y(i)=1$ represents the example is regarded as class $i$ by the classifier $f_i$, where $i\in \{1,\,2,\,\cdots,\,N\}$. When $sum(y)>1$, $x$ is classified as positives by more than one classifiers. Take Fig~\ref{fig:pipeline} for instance, after the beginning $N$ classifiers, $x$ is judged by $f_13$ firstly to determine whether $x$ belongs to class $1$. After then, the number of $1$ in $y^{(1)}$ is one less than that in $y$. Repeat input to other classifier $f_{ij}$ where $y(i)=y(j)=1$ until $sum(y)=1$. Finally, the class of $x$ determined as $c,\ y(c)=1\ and\ sum(y)=1$. If the initial $y$ has no $1$, the class of the related $x$ would be assigned randomly.

Here, we discuss the accuracy of this structure. Assuming that the accuracy of $f_i,\,i\in\{1,\,2,\,\cdots,\,N\}$ is $p_{accuracy}$, the probability of that one example is correctly classified by the supposed binary-classifier but mis-classified by other $K$ binary-classifiers are shown as equation~\ref{equ:probability wrong}. The output of the ensemble module would be right only under this condition. $K$ could be 0 and it means that the example is correctly classified.

\begin{equation}
\label{equ:probability wrong}
p_{K}=C_{N-1}^{K} \cdot p_{accuracy}^{N-K} \cdot (1-p_{accuracy})^K
\end{equation}
Then the example is input to the ensemble module. Assuming the accuracy of each $f_{ij}$ is $p_assemble$ equally, only if results of all used classifiers $f_ij,\ f_i(x)\cdot f_j(x)=1$ in the module are right, the example would be classified correctly. Taking no account of the random assignment, the probability of that the result of the multi-classifier is correct is shown as equation~\ref{equ:probability}.

\begin{equation}
\label{equ:probability}
p_{correct} = \sum\limits_{K=0}^{N-1}p_K \cdot p_{assemble}^{K}
\end{equation}

\begin{figure}
\begin{center}
\includegraphics[width=\textwidth]{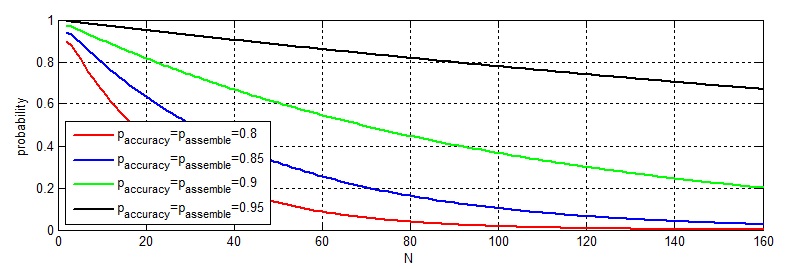}
\caption{the relationship between the probability of correct classification and $N$.}
\label{fig:probability correct}
\end{center}
\end{figure}

Fig~\ref{fig:probability correct} shows the relationship between the $p_{correct}$ and the number of class $N$. With the increase of $N$, the probability of correct classification is decreasing. As a result, this structure would work well when the number of classes are not too many. In contrast, it would perform terribly if this structure is used to classify the examples from a data including too many classes. Usually, if the data contains less than 20 classes, the above ensemble strategy is a satisfactory solution to synthesize these multiple BPT-trained binary-classifiers.

\section{Recognition Implementation}
\label{sec:recogition implementation}
This section introduces the specific implementation of the multi-classifier in section~\ref{sec:multi-bpt} for object recognition. It includes three parts: feature extraction, basic classifier and the implementation of "teachers".

The feature extraction is based on K-means~\cite{Coates2012Learning}, an unsupervised learning method. Support vector machine (SVM)~\cite{suykens1999least} is chosen as the basic classifier for BPT-trained binary-classifier. As for "teachers", we use logistic regression as the classification model, Histogram of Oriented Gradient(HOG)~\cite{Dalal2005Histograms} as the feature extraction method, and principal component analysis (PCA)~\cite{Jolliffe2010Principal} to decrease the dimensionality. Both the classification model and the feature extraction method are different from those of the basic classifier. These differences could improve the diversity.

Some labeled examples are reserved to train these "teachers". Every basic classifier needs two "teachers" to train itself. Furthermore, each classifier in the ensemble module is also need two "teachers". The two "teachers" are composed by one positive "teacher" and one negative "teacher". The positive "teacher" is to pick out false negative examples from the negative outputs of the binary-classifier. The negative "teacher" is to pick out false positive examples from the positive outputs. These "teachers" are trained by some examples randomly selected from these reserved labeled data. And the "teachers" are frequently re-trained in the proceeding of BPT, which aims to make sure that the "teachers" satisfy the equation~\ref{equ:condition}. In the process of re-training, the training data are re-selected randomly from the reserved labeled data. Random selection aims to avoid "teachers" being over-fitting and to improve the generalization of the model.

\begin{figure*}[htb]
\centering
\subfigure[negative "teachers"]{\includegraphics[width=0.4\textwidth]{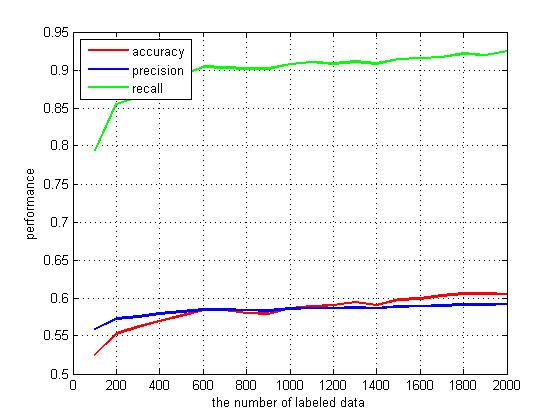}}
\subfigure[positive "teachers"]{\includegraphics[width=0.4\textwidth]{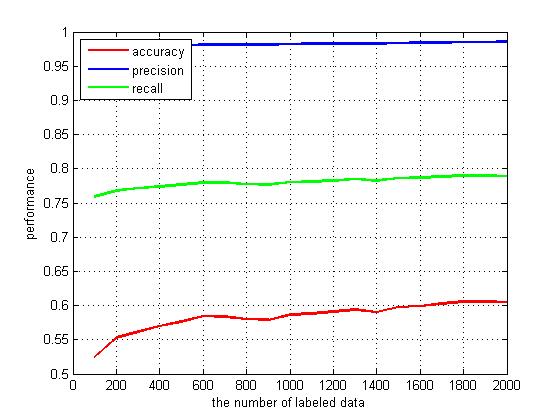}}
\caption{the average performances of negative and positive "teachers" trained in various data ranging from 100 to 2000.}
\label{fig:teachersperformance}
\end{figure*}

It is fine that only a few labeled data are used to train these "teachers" because two weak classifiers could train a strong classifier according to Liu's work~\cite{BoostPicking}. Fig~\ref{fig:teachersperformance} shows the average performance of positive "teachers" and negative "teachers" trained by various numbers of labeled data. The number of training data ranges from 100 to 2000. Accuracy, precision and recall of these "teachers" are recorded. Based on Fig~\ref{fig:teachersperformance}, it is concluded that the "teachers" could satisfy equation~\ref{equ:condition} when they are trained by a few labeled data.

\section{Experiment}
\label{sec:experiment}
This section details the experiment about object recognition to test the performance of BPT multi-classifier. In the proceeding of training the multi-classifier, only a few labeled data (compared to the size of the whole data) are used. And besides the labeled data, the rest data in the data set are also used as unlabeled data. We use CIFAR-10 and CIFAR-100~\cite{cifar} in the experiment and record the "accuracy" as the index of performance. In addition, we also compare the performance of the proposed method with those of previous and related methods.

In test, only 1000 labeled data are reserved to train these "teachers" in BPT. The rest of the training data (49000 totally) is used as unlabeled data. And the test set includes 10000 images. Table~\ref{table:cifar10} shows the performances of the proposed method in CIFIA-10 and CIFIA-100. Besides, there are some other methods~\cite{krizhevsky2012imagenet,residualnetwork,PCANet,SCNN,kmeannetwork,ckn,FMP,NIPS2014DUFL,NIPS2012_4824,bodnn,RecurrentCNN} compared with the proposed one in CIFIA-10. And we compare the proposed method with some other  methods~\cite{krizhevsky2012imagenet,highwaynetwork,pmu,NIN,DRL,OMP,ActivationDNN,ELUs,SPCNN,FMP,RecurrentCNN} in CIFIA-100. It should be noted that the training data of other methods are 100\% labeled, while the proposed method only use 2\% labeled data and 98\% unlabeled data in training set.

\begin{table}
\label{table:cifar10}
\begin{center}
\caption{Performance of various methods in CIFAR-10 and CIFAR-100}
\label{table:cifar10}
\begin{tabular}{llllll}
\hline\noalign{\smallskip}
 & &Tested in CIFAR-10& & & \\
\noalign{\smallskip}
\hline
\noalign{\smallskip}
method & accuracy & method & accuracy & method & accuracy\\
\noalign{\smallskip}
\hline
\noalign{\smallskip}
AlexNet~\cite{krizhevsky2012imagenet} & 81.96\% & \textbf{proposed method} & \textbf{78.39}\% & DUFL~\cite{NIPS2014DUFL} & 82 \% \\
PCANet~\cite{PCANet} & 78.67\% & S-CNN~\cite{SCNN} &75.96\% & CNN~\cite{NIPS2012_4824} & 89\%\\
K-means Net~\cite{kmeannetwork} & 77.4\% & Res-Net~\cite{residualnetwork} & 93.57\% & SBO~\cite{bodnn} & 96.63\% \\
CK-Net~\cite{ckn} &  82.18\% & FMP~\cite{FMP} & 96.53\% & Rec-CNN~\cite{RecurrentCNN} & 92.91\% \\
\hline
 & &Tested in CIFAR-100& & & \\
\noalign{\smallskip}
\hline
\noalign{\smallskip}
method & accuracy & method & accuracy & method & accuracy\\
\noalign{\smallskip}
\hline
\noalign{\smallskip}
AlexNet~\cite{krizhevsky2012imagenet} & 54.2\% & \textbf{proposed method} & \textbf{50.77}\% & PMU~\cite{pmu} & 61.86 \% \\
NIN~\cite{NIN} & 64.32\% & DRL~\cite{DRL} & 64.77\% & OMP~\cite{OMP} & 60.8\%\\
AF~\cite{ActivationDNN} & 69.17\% & ELUs~\cite{ELUs} & 75.72\% & SP-CNN~\cite{SPCNN} & 68.40\% \\
HighNet~\cite{highwaynetwork} &  67.76\% & FMP~\cite{FMP} & 73.61\% & Rec-CNN~\cite{RecurrentCNN} & 68.25\% \\
\hline
\end{tabular}
\end{center}
\end{table}
\setlength{\tabcolsep}{3pt}


In this experiment, the performances of the proposed method could almost approach the baseline, AlexNet~\cite{krizhevsky2012imagenet}. Especially in CIFAR-10, the proposed method performs even better than some supervised methods. In consideration of that the proposed method uses 2\% labeled data and 98\% unlabeled data, the performance of the proposed method is acceptable and reasonable. However, in CIFAR-100, the proposed method does not performs as good as it does in CIFAR-10. The main reason is that the ensemble strategy is not fit for the situation that the data includes too many classes, which is theoretically analyzed in section~\ref{subect:ensemble}.

\section{Conclusions}

The paper researches in object recognition based on a few labeled data and amounts of unlabeled data. Different from previous works on object recognition based on semi-supervised methods or weak-supervised methods, the key idea of the proposed method is that a supervised classification model could be trained by amounts of unlabeled data with their estimated labels as well as the same model trained by all labeled data. To achieve this goal, we design Boost Picking Teaching for binary-classification. Kalal's P-N learning~\cite{kalal2010pn,kalal2012tracking} and Liu's Boost Picking~\cite{BoostPicking} are the theoretical basises of BPT. Furthermore, how to expend binary-classification to multi-classification is discussed. This paper constructs an multi-classifier by synthesizing multiple binary-classifiers and analyzes the reason. In two object recognition data sets, BPT multi-classifier performs well even using just a few labeled data and amounts of unlabeled data. Especially when the number of class is less than dozens, the proposed model works effectively.

But there are two problems need further research. One is to design an ensemble method that is fit for many-class classification. Another one is improving the efficiency. BPT involves a lot of processes about re-training a classifier. The multi-classifier contains many BPT-trained binary-classifiers and every binary-classifier need a lot time to train well. The second one restrict the development of the proposed method seriously.


\clearpage

\bibliography{example_paper}

\begin{thebibliography}{48}
\providecommand{\natexlab}[1]{#1}
\providecommand{\url}[1]{\texttt{#1}}
\expandafter\ifx\csname urlstyle\endcsname\relax
  \providecommand{\doi}[1]{doi: #1}\else
  \providecommand{\doi}{doi: \begingroup \urlstyle{rm}\Url}\fi

\bibitem[ELU()]{ELUs}
Fast and accurate deep network learning by exponential linear units (elus).

\bibitem[Alexey~Dosovitskiy \& Brox(2014)Alexey~Dosovitskiy and
  Brox]{NIPS2014DUFL}
Alexey~Dosovitskiy, Jost Tobias~Springenberg, Martin~Riedmiller and Brox,
  Thomas.
\newblock Discriminative unsupervised feature learning with convolutional
  neural networks.
\newblock NIPS, 2014.

\bibitem[Attenberg et~al.(2011)Attenberg, Ipeirotis, and
  Provost]{attenberg2011beat}
Attenberg, Josh, Ipeirotis, Panagiotis~G, and Provost, Foster~J.
\newblock Beat the machine: Challenging workers to find the unknown unknowns.
\newblock \emph{Human Computation}, 11:\penalty0 11, 2011.

\bibitem[Bertsekas(2015)]{Bertsekas2015Incremental}
Bertsekas, Dimitri~P.
\newblock Incremental gradient, subgradient, and proximal methods for convex
  optimization: A survey.
\newblock \emph{Optimization}, 2015.

\bibitem[Brooms(2006)]{Brooms2006Stochastic}
Brooms, A.~C.
\newblock Stochastic approximation and recursive algorithms with applications,
  2nd edn by h. j. kushner and g. g. yin.
\newblock \emph{Journal of the Royal Statistical Society}, 169\penalty0
  (3):\penalty0 654--654, 2006.

\bibitem[Burton et~al.(2012)Burton, Brady, Brewer, Neylan, Bigham, and
  Hurst]{burton2012crowdsourcing}
Burton, Michele~A, Brady, Erin, Brewer, Robin, Neylan, Callie, Bigham,
  Jeffrey~P, and Hurst, Amy.
\newblock Crowdsourcing subjective fashion advice using vizwiz: challenges and
  opportunities.
\newblock In \emph{Proceedings of the 14th international ACM SIGACCESS
  conference on Computers and accessibility}, pp.\  135--142. ACM, 2012.

\bibitem[Chen et~al.(2013)Chen, Shrivastava, and Gupta]{Chen2013NEIL}
Chen, Xinlei, Shrivastava, A., and Gupta, A.
\newblock Neil: Extracting visual knowledge from web data.
\newblock In \emph{Computer Vision, IEEE International Conference on}, pp.\
  1409--1416, 2013.

\bibitem[Coates \& Ng.(2011)Coates and Ng.]{kmeannetwork}
Coates, A. and Ng., A.~Y.
\newblock Selecting receptive fields in deep networks.
\newblock pp.\  2528¨C2536. NIPS, 2011.

\bibitem[Coates \& Ng(2012)Coates and Ng]{Coates2012Learning}
Coates, Adam and Ng, Andrew~Y.
\newblock Learning feature representations with k-means.
\newblock \emph{Lecture Notes in Computer Science}, 7700:\penalty0 561--580,
  2012.

\bibitem[Dalal \& Triggs(2005)Dalal and Triggs]{Dalal2005Histograms}
Dalal, Navneet and Triggs, Bill.
\newblock Histograms of oriented gradients for human detection.
\newblock In \emph{IEEE Conference on Computer Vision \& Pattern Recognition},
  pp.\  886--893, 2005.

\bibitem[Fei-Fei(2010)]{fei2010imagenet}
Fei-Fei, L.
\newblock Imagenet: crowdsourcing, benchmarking \& other cool things.
\newblock In \emph{CMU VASC Seminar}, 2010.

\bibitem[Felzenszwalb et~al.(2010)Felzenszwalb, Girshick, David, and
  Deva]{Felzenszwalb2010Object}
Felzenszwalb, Pedro~F, Girshick, Ross~B, David, Mc~Allester, and Deva, Ramanan.
\newblock Object detection with discriminatively trained part-based models.
\newblock \emph{Pattern Analysis and Machine Intelligence IEEE Transactions
  on}, 32\penalty0 (9):\penalty0 1627--1645, 2010.

\bibitem[Fergus et~al.(2009)Fergus, Weiss, and Torralba]{Fergus2009Semi}
Fergus, Rob, Weiss, Yair, and Torralba, Antonio.
\newblock Semi-supervised learning in gigantic image collections.
\newblock In \emph{Advances in Neural Information Processing Systems 22: 23rd
  Annual Conference on Neural Information Processing Systems 2009. Proceedings
  of a meeting held 7-10 December 2009, Vancouver, British Columbia, Canada.},
  2009.

\bibitem[Forest~Agostinelli(2015)]{ActivationDNN}
Forest~Agostinelli, Matthew~Hoffman, Peter Sadowski Pierre~Baldi.
\newblock Learning activation functions to improve deep neural networks.
\newblock CVPR, 2015.

\bibitem[Fuqiang~Liu(2016)]{BoostPicking}
Fuqiang~Liu, Fukun~Bi, Yiding Yang Liang~Chen.
\newblock Boost picking: A novel method on converting supervised classification
  to semi-supervised classification.
\newblock 2016.

\bibitem[Geman et~al.(1992)Geman, Bienenstock, and Doursat]{geman1992neural}
Geman, Stuart, Bienenstock, Elie, and Doursat, Ren{\'e}.
\newblock Neural networks and the bias/variance dilemma.
\newblock \emph{Neural computation}, 4\penalty0 (1):\penalty0 1--58, 1992.

\bibitem[Graham(2015)]{FMP}
Graham, Benjamin.
\newblock Fractional max-pooling.
\newblock 2015.

\bibitem[Hosmer~Jr \& Lemeshow(2004)Hosmer~Jr and Lemeshow]{hosmer2004applied}
Hosmer~Jr, David~W and Lemeshow, Stanley.
\newblock \emph{Applied logistic regression}.
\newblock John Wiley \& Sons, 2004.

\bibitem[James(2003)]{james2003variance}
James, Gareth~M.
\newblock Variance and bias for general loss functions.
\newblock \emph{Machine Learning}, 51\penalty0 (2):\penalty0 115--135, 2003.

\bibitem[Jasper~Snoek(2015)]{bodnn}
Jasper~Snoek, Oren~Rippel, Kevin Swersky Ryan Kiros Nadathur Satish Narayanan
  Sundaram Md. Mostofa Ali Patwary Prabhat Ryan P.~Adams.
\newblock Scalable bayesian optimization using deep neural networks.
\newblock ICML, 2015.

\bibitem[Jolliffe(2010)]{Jolliffe2010Principal}
Jolliffe, I.~T.
\newblock Principal component analysis.
\newblock \emph{Springer Berlin}, 87\penalty0 (100):\penalty0 41--64, 2010.

\bibitem[Jost Tobias~Springenberg(2013)]{pmu}
Jost Tobias~Springenberg, Martin~Riedmiller.
\newblock Improving deep neural networks with probabilistic maxout units.
\newblock ICML, 2013.

\bibitem[Julien~Mairal(2014)]{ckn}
Julien~Mairal, Piotr~Koniusz, Zaid~Harchaoui.
\newblock Convolutional kernel networks.
\newblock 2014.

\bibitem[K.~Zhou \& Glover(1996)K.~Zhou and Glover]{robustandoptimalcontrol}
K.~Zhou, J.C.~Doyle and Glover, K.
\newblock Robust and optimal control.
\newblock 1996.

\bibitem[Kaiming~He(2015)]{residualnetwork}
Kaiming~He, Xiangyu~Zhang, Shaoqing Ren Jian~Sun.
\newblock Deep residual learning for image recognition.
\newblock 2015.

\bibitem[Kalal et~al.(2010)Kalal, Matas, and Mikolajczyk]{kalal2010pn}
Kalal, Zdenek, Matas, Jiri, and Mikolajczyk, Krystian.
\newblock Pn learning: Bootstrapping binary classifiers by structural
  constraints.
\newblock In \emph{Computer Vision and Pattern Recognition (CVPR), 2010 IEEE
  Conference on}, pp.\  49--56. IEEE, 2010.

\bibitem[Kalal et~al.(2012)Kalal, Mikolajczyk, and Matas]{kalal2012tracking}
Kalal, Zdenek, Mikolajczyk, Krystian, and Matas, Jiri.
\newblock Tracking-learning-detection.
\newblock \emph{Pattern Analysis and Machine Intelligence, IEEE Transactions
  on}, 34\penalty0 (7):\penalty0 1409--1422, 2012.

\bibitem[Krizhevsky(2009)]{cifar}
Krizhevsky, Alex.
\newblock Learning multiple layers of features from tiny images.
\newblock 2009.

\bibitem[Krizhevsky et~al.(2012{\natexlab{a}})Krizhevsky, Sutskever, and
  Hinton]{NIPS2012_4824}
Krizhevsky, Alex, Sutskever, Ilya, and Hinton, Geoffrey~E.
\newblock Imagenet classification with deep convolutional neural networks.
\newblock pp.\  1097--1105. NIPS, 2012{\natexlab{a}}.

\bibitem[Krizhevsky et~al.(2012{\natexlab{b}})Krizhevsky, Sutskever, and
  Hinton]{krizhevsky2012imagenet}
Krizhevsky, Alex, Sutskever, Ilya, and Hinton, Geoffrey~E.
\newblock Imagenet classification with deep convolutional neural networks.
\newblock In \emph{Advances in neural information processing systems}, pp.\
  1097--1105, 2012{\natexlab{b}}.

\bibitem[Li et~al.(2006)Li, Rob, and Pietro]{Li2006One}
Li, Fei~Fei, Rob, Fergus, and Pietro, Perona.
\newblock One-shot learning of object categories.
\newblock \emph{IEEE Transactions on Pattern Analysis \& Machine Intelligence},
  28\penalty0 (4):\penalty0 594--611, 2006.

\bibitem[Liang et~al.(2014)Liang, Liu, Wei, Liu, Lin, and
  Yan]{Liang2014Computational}
Liang, Xiaodan, Liu, Si, Wei, Yunchao, Liu, Luoqi, Lin, Liang, and Yan,
  Shuicheng.
\newblock Computational baby learning.
\newblock \emph{Eprint Arxiv}, 2014.

\bibitem[Lin et~al.(2014)Lin, Maire, Belongie, Hays, Perona, Ramanan,
  Doll{\'a}r, and Zitnick]{lin2014microsoft}
Lin, Tsung-Yi, Maire, Michael, Belongie, Serge, Hays, James, Perona, Pietro,
  Ramanan, Deva, Doll{\'a}r, Piotr, and Zitnick, C~Lawrence.
\newblock Microsoft coco: Common objects in context.
\newblock In \emph{Computer Vision--ECCV 2014}, pp.\  740--755. Springer, 2014.

\bibitem[Mark D.~McDonnell(2015)]{SCNN}
Mark D.~McDonnell, Tony~Vladusich.
\newblock Enhanced image classification with a fast-learning shallow
  convolutional neural network.
\newblock 2015.

\bibitem[Min~Lin()]{NIN}
Min~Lin, Qiang~Chen, Shuicheng~Yan.
\newblock Network in network.

\bibitem[Ming~Liang()]{RecurrentCNN}
Ming~Liang, Xiaolin~Hu.
\newblock Recurrent convolutional neural network for object recognition.

\bibitem[Mohri et~al.(2012)Mohri, Rostamizadeh, and
  Talwalkar]{mohri2012foundations}
Mohri, Mehryar, Rostamizadeh, Afshin, and Talwalkar, Ameet.
\newblock \emph{Foundations of machine learning}.
\newblock MIT press, 2012.

\bibitem[Oren~Rippel(2015)]{SPCNN}
Oren~Rippel, Jasper~Snoek.
\newblock Spectral representations for convolutional neural networks.
\newblock NIPS, 2015.

\bibitem[Prest et~al.(2012)Prest, Leistner, Civera, and
  Schmid]{Prest2012Learning}
Prest, A., Leistner, C., Civera, J., and Schmid, C.
\newblock Learning object class detectors from weakly annotated video.
\newblock In \emph{IEEE Conference on Computer Vision \& Pattern Recognition},
  pp.\  3282--3289, 2012.

\bibitem[Rosenberg et~al.(2005)Rosenberg, Hebert, and
  Schneiderman]{Rosenberg2005Semi}
Rosenberg, Chuck, Hebert, Martial, and Schneiderman, Henry.
\newblock Semi-supervised self-training of object detection models.
\newblock In \emph{7th IEEE Workshop on Applications of Computer Vision / IEEE
  Workshop on Motion and Video Computing (WACV/MOTION 2005), 5-7 January 2005,
  Breckenridge, CO, USA}, pp.\  29--36, 2005.

\bibitem[Rupesh Kumar~Srivastava(2015)]{highwaynetwork}
Rupesh Kumar~Srivastava, Klaus~Greff, Jurgen~Schmidhuber.
\newblock Highway networks.
\newblock 2015.

\bibitem[Shalev-Shwartz(2012)]{Shalev2012Online}
Shalev-Shwartz, Shai.
\newblock Online learning and online convex optimization.
\newblock \emph{Foundations and Trends in Machine Learning}, 4\penalty0
  (2):\penalty0 107--194, 2012.

\bibitem[Shuo~Yang \& Tang(2015)Shuo~Yang and Tang]{DRL}
Shuo~Yang, Ping~Luo, Chen Change Loy Kenneth W.~Shum1 and Tang, Xiaoou.
\newblock Deep representation learning with target coding.
\newblock AAAI, 2015.

\bibitem[Socher et~al.(2013)Socher, Ganjoo, Sridhar, Bastani, Manning, and
  Ng]{Socher2013Zero}
Socher, Richard, Ganjoo, Milind, Sridhar, Hamsa, Bastani, Osbert, Manning,
  Christopher~D., and Ng, Andrew~Y.
\newblock Zero-shot learning through cross-modal transfer.
\newblock \emph{Advances in Neural Information Processing Systems}, pp.\
  935--943, 2013.

\bibitem[Sung \& Poggio(1995)Sung and Poggio]{Sung1995Example}
Sung, Kah~Kay and Poggio, Tomaso.
\newblock Example based learning for view-based human face detection.
\newblock \emph{Pattern Analysis and Machine Intelligence IEEE Transactions
  on}, 20\penalty0 (1):\penalty0 39--51, 1995.

\bibitem[Suykens \& Vandewalle(1999)Suykens and Vandewalle]{suykens1999least}
Suykens, Johan~AK and Vandewalle, Joos.
\newblock Least squares support vector machine classifiers.
\newblock \emph{Neural processing letters}, 9\penalty0 (3):\penalty0 293--300,
  1999.

\bibitem[Tsung-Han~Chan(2014)]{PCANet}
Tsung-Han~Chan, Kui~Jia, Shenghua Gao Jiwen Lu Zinan Zeng Yi~Ma.
\newblock Pcanet: A simple deep learning baseline for image classification?
\newblock 2014.

\bibitem[Tsung-Han~Lin(2014)]{OMP}
Tsung-Han~Lin, H. T.~Kung.
\newblock Stable and efficient representation learning with nonnegativity
  constraints.
\newblock ICML, 2014.

\end{thebibliography}
\bibliographystyle{icml2016}

\end{document}